\documentclass[letterpaper, 10 pt, conference]{ieeeconf}  

\IEEEoverridecommandlockouts                              
\usepackage[utf8]{inputenc} 
\usepackage[T1]{fontenc}
\usepackage{textcomp}
\usepackage{amsmath}
\usepackage[font=small, skip=0.1cm]{caption}
\usepackage[english]{babel}
\usepackage{booktabs}
\usepackage[nolist, nohyperlinks]{acronym}
\usepackage{subcaption}
\usepackage{tikz}
\usepackage{gensymb}

\title{EMBC conf}
\title{\LARGE \bf Continuous Non-Invasive Eye Tracking In Intensive Care}

\author{Ahmed Al-Hindawi$^{1,2}$, Marcela P Vizcaychipi$^{2,3}$ and Yiannis Demiris$^{1}$
\thanks{$^{1}$ Personal Robotics Laboratory, Department of Electrical and Electronic Engineering, Imperial College London, London SW7 2AZ, United Kingdom. {\tt\small a.al-hindawi@imperial.ac.uk}.}
\thanks{$^{2}$ Magill Department of Anaesthesia, Chelsea \& Westminster Hospital NHS Foundation Trust, London SW10 9NH, United Kingdom.} \\
\thanks{${3}$ Anaesthetics, Pain Medicine and Intensive Care, Department of Cancer \& Surgery, Imperial College London, London SW7 2AZ, United Kingdom.}}

\begin{document}

\begin{acronym}
    \acro{ROS}{Robot Operating System}
    \acro{AI}{Artificial Intelligence}
    \acro{JIA}{Juvenile Idiopathic Arthritis}
    \acro{CAM-ICU}{Confusion Assessment Method in ICU}
    \acro{CAM}{Confusion Assessment Method}
    \acro{APRV}{Airway Pressure Release Ventilation}
    \acro{ARDS}{Acute Respiratory Distress Syndrome}
    \acro{AUROC}{Area Under Receiver Operator Curve}
    \acro{BIS}{BiSpectral Index}
    \acro{CAP}{The Cyclic Alternating Pattern of EEG activity during sleep}
    \acro{CFP}{Corneo-Fundal Potential}
    \acro{CNN}{Convolutional Neural Network}
    \acro{CRP}{C-Reactive Protein}
    \acro{DeepMLNet}{Deep Multilayer Neural Network}
    \acro{dlPFC}{Dorso-Lateral Pre-Frontal Cortex}
    \acro{DMARD}{Disease Modifying Anti-rheumatic Drug}
    \acro{DSM}{Diagnostic and Statistical Manual of Mental Disorders}
    \acro{DTW}{Dynamic Time Warping}
    \acro{ECG}{Electrocardiograph}
    \acro{EEG}{Electroencephalogram}
    \acro{EOG}{Electro-oculograph}
    \acro{GBMR}{Graph Based Manifold Ranking}
    \acro{HAI}{Hospital Acquired Infection}
    \acro{HMM}{Hidden Markov Model}
    \acro{ICU}{Intensive Care Unit}
    \acro{LARS}{Least Angle Regression}
    \acro{LED}{Light Emitting Diode}
    \acro{MIMIC}{Multiparameter Intelligent Monitoring in Intensive Care}
    \acro{NICE}{National Institute of Clinical Excellence}
    \acro{PCBC}{Predictive Coding/Bias Competition}
    \acro{PDF}{Probability Density Function}
    \acro{PRE-DELIRIC}{PREdiction of DELIRium in ICu patients}
    \acro{REM}{Rapid Eye Movements}
    \acro{RF}{Receptive Field}
    \acro{ROC}{Receiver Operator Characteristic}
    \acro{ROI}{Regions Of Interest}
    \acro{RPE}{Retinal Pigment Epithelium}
    \acro{SAPSII}{Simplified Acute Physiology Score II}
    \acro{SC}{Sparse Coding}
    \acro{LARS}{Least Angle Regression}
    \acro{PCBC}{Predictive Coding/Bias Competition}
    \acro{AUROC}{Area Under Receiver Operator Curve}
    \acro{HAMMER}{Hierarchial Attentive Multiple Models for Execution and Recognition}
    \acro{LTSM}{Long Short-Term Memory}
    \acro{SVM}{Support Vector Machine}
    \acro{S$^3$FD}{Single Shot Scale-invariant Face Detector}
    \acro{3DDFA}{Face Alignment in Full Pose Range: A 3D Total Solution}
    \acro{RT-GENE}{Real-Time Eye Gaze Estimation in Natural Environments}
    \acro{RT-BENE}{Real-Time Blink Estimation in Natural Environments}
    \acro{RANSAC}{Random Sample Consensus}
    \acro{EMG}{Electromyogram}
    \acro{RGB}{Red-Green-Blue}
    \acro{RGBD}{Red-Green-Blue-Depth}
    \acro{MTCNN}{Multi-Task Cascaded Convolutional Networks for Joint Face Detection and Alignment}
    \acro{MSE}{Mean Square Error}
    \acro{HopeNet}{Fine-Grained Head Pose Estimation Without Keypoints}
\end{acronym}

\maketitle{}
\thispagestyle{empty}
\pagestyle{empty}

\def\requiredacc{15.4\degree~}
\def\requiredprec{6.4\degree~}
\def\evalacc{4.4\degree~}
\def\evalprec{2.7\degree~}
\begin{abstract}
Delirium, an acute confusional state, is a common occurrence in \acfp{ICU}. Patients who develop delirium have globally worse outcomes than those who do not and thus the diagnosis of delirium is of importance. Current diagnostic methods have several limitations leading to the suggestion of eye-tracking for its diagnosis through  \textit{in-attention}. To ascertain the requirements for an eye-tracking system in an adult \ac{ICU}, measurements were carried out at Chelsea \& Westminster Hospital NHS Foundation Trust. Clinical criteria guided empirical requirements of invasiveness and calibration methods while accuracy and precision were measured. A non-invasive system was then developed utilising a patient-facing \acs{RGB} camera and a scene-facing \acs{RGBD} camera. The system's performance was measured in a replicated laboratory environment with healthy volunteers revealing an accuracy and precision that outperforms what is required while simultaneously being non-invasive and calibration-free 
The system was then deployed as part CONfuSED, a clinical feasibility study where we report aggregated data from 5 patients as well as the acceptability of the system to bedside nursing staff. The system is the first eye-tracking system to be deployed in an \ac{ICU}.
\end{abstract}


\acresetall{}

\section{Introduction}

Delirium is an acute confusion state that is a fluctuating, usually reversible, cause of cerebral dysfunction that manifests clinically with a wide range of neuropsychiatric abnormalities. This state can occur in any acutely unwell patient but occurs with high incidence on \ac{ICU} owing to the acuity of diseases \cite{krewulakRiskFactorsOutcomes2020}. An estimate of the incidence of delirium in acutely unwell patients is 20\% and has been reported as high as 80\% \cite{linImpactDeliriumSurvival2004,pisaniResearchAlgorithmImprove2006}. 

The development of delirium in patients leads to increased risk of dying, increased hospital length of stay, increased cost of stay and worse cognitive scores compared to patients who do not \cite{clancyPsychologicalNeurocognitiveConsequences2015,mccuskerDeliriumPredicts12Month2002, linImpactDeliriumSurvival2004}. Thus, the contemporaneous diagnosis of delirium is of paramount importance. Current methods of diagnosis are laborious and often miss episodes of delirium due to the disease being time-variant, or are only applicable to a small subset of patients.


\begin{figure}[!t]
	\centering
    \includegraphics[width=\linewidth]{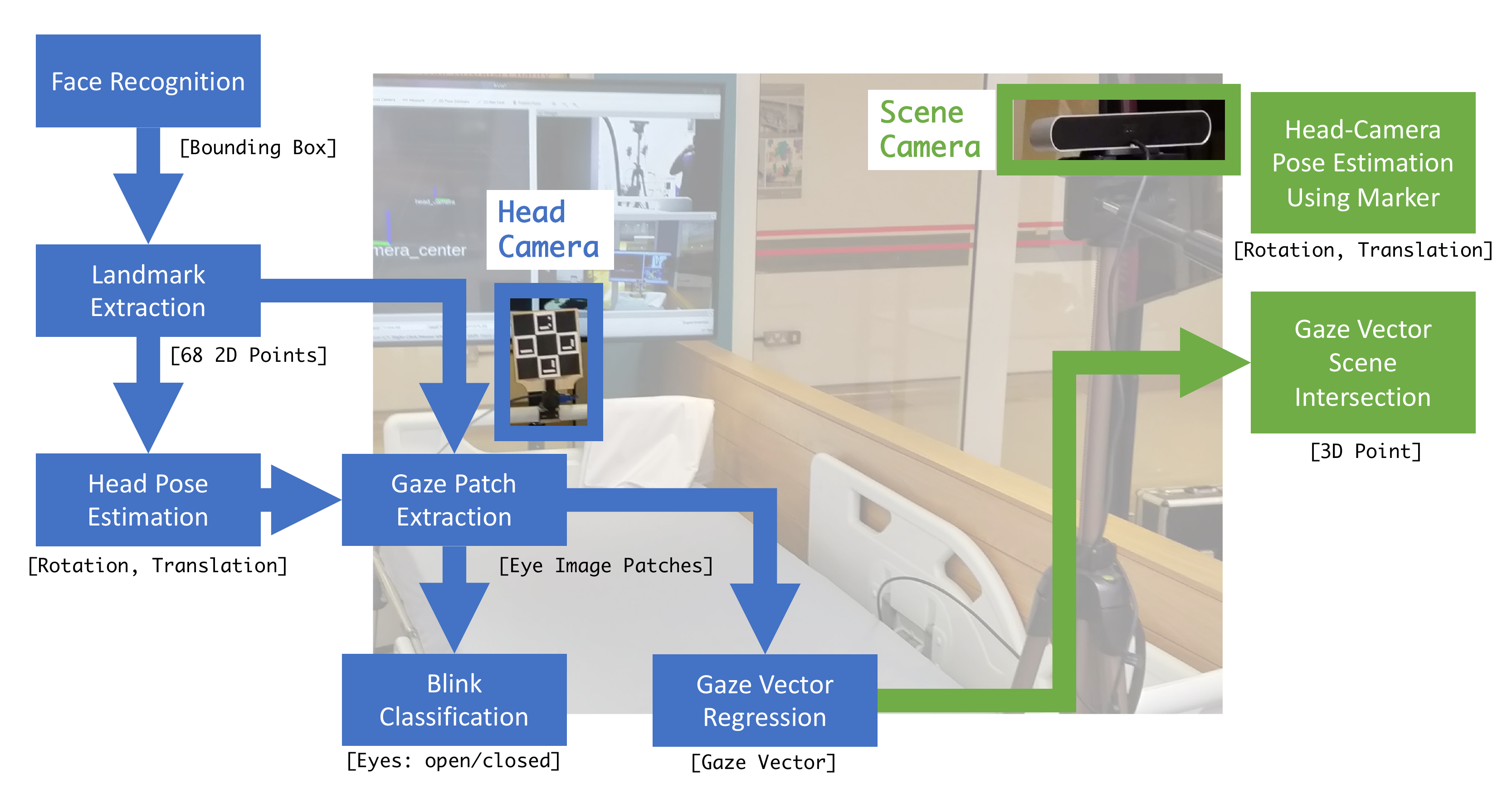}
	\caption{The proposed system for eye tracking in \ac{ICU}. Two cameras are used, an RGB patient facing camera (head-camera, blue) and a perspective facing RGBD camera (scene-camera, green) connected to a laptop (outside the viewpoint) running \ac{ROS}. The flow of information through the system is demonstrated with the head-camera and scene-camera running concurrently.}
	\label{fig:setup_flow}
\end{figure}

To aid in the timely diagnosis of delirium, we propose a non-invasive system that relies on eye-tracking as a biological signal of \textit{inattention} as a surrogate marker for delirium. The system is continuous, provides data in real-time, does not require patient involvement for calibration, is accurate and does not limit or restrict movements. Empiric measurements in an adult \ac{ICU} led to a set of requirements around invasiveness, calibration, accuracy, and precision. Under replicated laboratory conditions, the performance of the system was shown to be suitable for use in an adult \ac{ICU} where it outperforms requirements. Following ethical and governance approvals for a clinical feasibility study (CONfuSED \footnote{https://clinicaltrials.gov/ct2/show/NCT04589169}), the system was deployed on patients in \ac{ICU} to test its suitability. We report on descriptive statistical data on the initial 5 patients. We also report on the acceptability of the system to the bedside nursing staff.

\section{Related Work \& Diagnostic Methods}
Manual assessment methods, specifically the \ac{CAM-ICU}, are currently the most widely adopted systems in use for the diagnosis of delirium \cite{elyDeliriumMechanicallyVentilated2001,elyEvaluationDeliriumCritically2001}. The test relies on \textit{inattention} as its primary method of diagnosis of delirium. 

\acf{EEG} is of limited use owing to technological limitations but more recent attempts use deep neural networks for the diagnosis of delirium in a clinically representative population \cite{vanderkooiDeliriumDetectionUsing2015,koponenEEGSpectralAnalysis1989, sunAutomatedTrackingLevel2019}. \ac{EEG}, however, requires the placement of electrodes on a patient's scalp for a prolonged period to ensure adequate signal acquisition hampering its clinical safety. 

Eye-tracking technology has had success in inferring \textit{covert} internal mental states \cite{amadoriDecisionAnticipationDriving2020, wangRealTimeWorkloadClassification2018}. Clinically, eye-tracking is established in numerous fields where its success has led to the suggestion that it can be used for the diagnosis of delirium through \textit{inattention} \cite{trillenbergEyeMovementsPsychiatric2004,extonEyeTrackingTechnology2009}. The application of eye-tracking in \ac{ICU} has been limited by the technological limitations of currently available systems. To the knowledge of the authors, no suitable system has been developed or implemented that facilitates the acquisition of eye movements from delirious patients in a clinical setting. 

\section{System Requirements}
\label{sec:system_requirements}
\ac{ICU} presents a challenge for eye-tracking technology. The mixture of patient, disease, environment, staff and medical equipment regulations all coalesce together placing the following clinical empirical requirements:
\paragraph{Non-Invasive} 
Due to the nature of patients and their disease in \ac{ICU}, a non-invasive system is required owing to the need for clinical hands-on care. Having a device that instruments the patient will interfere with clinical care and is thus potentially unsafe. This also infers that a device that facilitates free-viewing is required -- a system that does not restrict the patient for signal acquisition.
\paragraph{Calibration-Free}
In delirium, patients are unable to follow commands making calibration through command-following difficult. The system must be able to track the patient's gaze without an explicit calibration step that maps eye movements to scene gaze positions.
\paragraph{Accurate}
The required accuracy of a system is dependent on its domain use; In \ac{ICU}, accuracy facilitates the identification of \textit{which} object the patient is looking at and correlates with the \textit{spread} of objects in the patient's perspective. 
\paragraph{Precise}
Precision, defined as the reproducibility of measurement, is important as low precision leads to uncertainty on the gaze estimate. Low uncertainty facilitates subsequent analysis through the reliable classification of eye movements. In this context, the \textit{size} of objects from the point of view of the patient correlates with precision. High precision leads to reproducible measurements which leads to low spread and the ability to identify small objects.
\paragraph{Performant}
Eye movement research suggests that fixations typically last between 0.1 - 0.5 seconds giving a minimum required frequency of 20Hz by Nyquist criteria \cite{fengEyeMovementsTimeseries2006}. This system would ideally run on commercially available hardware.

\section{Methodology}
To meet the challenges set out by the empirical requirements, we have designed a system that decouples the regression of the patient's eye and head pose from where they are looking in the scene through the use of a dual-camera system as per Figure~\ref{fig:setup_flow}. One camera located in front of the patient at the foot end of the bed facing the patient (termed the head-camera) and another behind the patient facing the same direction as the patient (termed the scene camera). The head-camera is a \ac{RGB} camera whereas the scene-camera is required to be a \ac{RGBD} camera. The decoupling of the eye gaze regression and gaze inference through the use of two cameras satisfies the first requirement of \textit{Non-invasiveness} as per Section~\ref{sec:system_requirements}.

\paragraph{Scene-Camera}
\label{subsec:scene_camera}
The scene-camera performs head-camera pose estimation and gaze vector-scene intersection. The head camera's pose is measured using a ChAruCo board which is fixed superior to the head-camera -- a once only calibration procedure yields a static pose between the head-camera and its board \cite{anCharucoBoardBasedOmnidirectional2018}.
The resolution of the scene camera is dependent on its ability to resolve this marker reliably - we utilise a 1920x1080 pixel resolution at 30Hz.
Once the gaze-vector has been estimated from the head camera, it is the scene camera that resolves that vector as a point in the scene. This is done using a ray-octree intersection that is created from the point cloud received from the scene camera. \cite{glassnerSpaceSubdivisionFast1984}.

\begin{figure}[t!]
    \centering{}
    \input{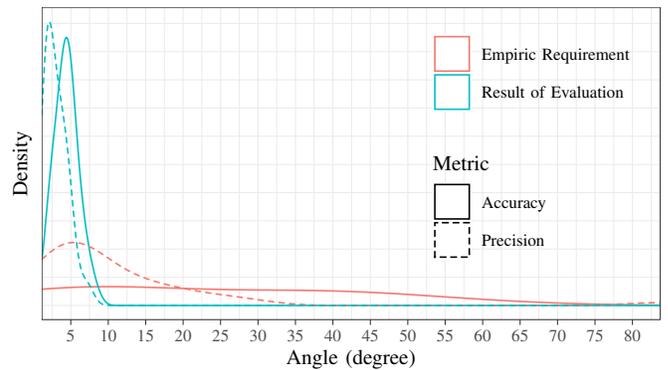}
    \caption{Density plots demonstrating the required accuracy and precision from empirical measurements and the system's metrics as evaluated in the replicated environment in the laboratory.}
    \label{fig:requirement_evaluation}
\end{figure}

\paragraph{Head-Camera}
\label{para:head_camera}
The patient-facing \textit{head-camera} carries out multiple steps that ultimately result in the head-pose and gaze-vector of the patient relative to the head camera. The head camera's pose as estimated from the scene camera creates a transformation tree that facilitates the location of the patient's gaze relative to the scene camera which acquires the point of view of the patient.

The first step is face detection. Following experimental evaluation between various face detectors, the \ac{S$^3$FD} face detector was used for its robust detection at extreme head poses \cite{zhangFDSingleShot2017}. The output is a bounding box around potential faces and the confidence of each detection. The camera specification is linked to its ability to reliably produce a high-resolution image of the face - we utilise a 1024x768 pixel resolution at 50Hz.

Following face-detection, the patient's head-pose is then estimated by solving a perspective-n-point problem that aims to align a generic 3D model of a human's head with the extracted landmarks from the patient \cite{shayohayonRobust3DHead2006}. \ac{MTCNN}, \ac{HopeNet}, and \ac{3DDFA} were evaluated for head-pose estimation using a composite score comprised of accuracy, performance at inference time and utilisation of GPU Memory. \ac{3DDFA} had the lowest composite score and hence was chosen. Its output is 68 2D landmark points in image space that consistently outline the facial features of the patient's face. By knowing the camera calibration matrix and having a predefined model, a direct least squares method is utilised following a \ac{RANSAC} scheme to obtain the pose of the patient's head relative to the head-camera \cite{heschDirectLeastSquaresDLS2011, fischlerRandomSampleConsensus1981}. The 3D model's size is parametrised by the patient's inter-pupillary distance with 0.06 metres used as the default.
\begin{table}[t]
    \centering
    \begin{tabular}{p{0.5\linewidth}p{0.2\linewidth}p{0.2\linewidth}}
    \toprule
    Variable                       & Validation (n=6) & CONfuSED (n=5) \\ \midrule
    Age                            & 25 - 32          & 18 - 73 \\
    Gender (male = 1)              & 5                & 3\\
    Vision Corrected (Glasses = 1) & 2                & 2\\ \bottomrule 
    \end{tabular}
    \caption{Demographics data of participants in validation experiment and patients in CONfuSED.}
    \label{tab:demographics}
\end{table}
The landmarks extracted from the previous stage also result in the extraction of two image patches of the patient's eyes which are then used for gaze regression. \ac{RT-GENE}, a neural network-based model that regresses gaze vector from eye patches, was chosen for its published accuracy in natural viewing environments while \ac{RT-BENE} is used for blink detection following improvement through a data augmentation scheme and the use of a ResNet-18 as the backbone for increased speed and accuracy \cite{fischerRTGENERealTimeEye2018, cortaceroRTBENEDatasetBaselines2019}. \ac{RT-GENE} was modified in this work to optimise inter-process communication and reduce latency between image capture and gaze estimation. The resolution of the camera (1024x768) results in eye-patches that are down-sampled for the consumption through \ac{RT-GENE}/\ac{RT-BENE} at a clinically safe distance of approximately 2~metres. The use of \ac{RT-GENE} and \ac{RT-BENE} in this pipeline satisfies the requirement for the system being \textit{Calibration-Free} as per Section~\ref{sec:system_requirements}.

The required accuracy and precision from an eye tracker in \ac{ICU} were measured in situ. Measurements of the scene from the point of view of 5 patient from 5 different environments were undertaken across our \ac{ICU}. Accuracy, as defined as the difference between a target location and the measured gaze relative to the user's head correlates with the identification of \textit{which} object in the scene the patient is looking at and thus, given a collection of objects in the scene, the average \textit{distance} between objects is of interest. The distances between objects in the viewpoint of the patient and their relative distance to the patient's head were measured. The average of these measurements then form the minimum required accuracy of the system.
To ascertain a similar requirement for precision, the size of objects and their distance to the patient's head was measured. In this context, the size of the object in the scene correlates to repeatability -- the smaller the size of the object, the higher the precision required for the identification of the object the patient is looking at. The average of these measurements then forms the required precision. Formally:

\begin{align}
    \text{Accuracy} &= \frac{1}{N}\sum_{i=1}^{N}{cos^{-1}(|gaze_{i}| \cdot |target|)} \label{eq:accuracy}\\
    \text{Precision} &= \sqrt{\frac{1}{N}\sum_{i=1}^{N-1}{cos^{-1}(|gaze_{i+1}| \cdot |gaze_{i}|)^{2}}} \label{eq:precision}
\end{align}
where $\cdot$ is the vector dot product and $|x|$ is vector normalisation to unit length. The gaze vector is measured to start from the head pose's transform and extends to a unit length. The target vector is measured to also start from the head pose transform and extends to the target.

To measure the performance of the system, six healthy, vision-corrected, non-delirious volunteers were recruited to undergo testing of the accuracy and precision of the system (Table~\ref{tab:demographics}). An \ac{ICU} replicated environment was created under laboratory conditions and the bed was inclined at 30 degrees. The participant's head-pose and gaze vector were regressed from the head camera placed on the bottom left-hand side of the screen. Each participant was asked to look at the centre of an AruCo marker that would appear at a random location of the screen that was 0.1~metres in size; 40 samples were taken per marker. The experimental endpoint was defined as full coverage of the screen by target markers.
For each marker location, outliers consisting of the first 10 and last 10 measurements were discarded concentrating on the middle 20 measurements to remove gaze points leading into and out of the target marker. For each marker location, the accuracy and precision were calculated as per Equations~\ref{eq:accuracy}~\&~\ref{eq:precision}. For visualisation, median accuracy for each marker is then rasterised onto an image with the resolution of the screen which is then smoothed with a Gaussian kernel.

As part of a clinical feasibility pilot trial that aims to look at at the use of this system on the detection of delirium in \ac{ICU}, the system was deployed on 5 patients. We report aggregated descriptive statistical plots to demonstrate the performance of the system in patients suffering from episodes of delirium compared to episodes without delirium.

In assessing the acceptability of the system clinically, a questionnaire was deployed to the staff that were nursing the patients recruited into CONfuSED (Table~\ref{tab:nurse_questionnaire}) \cite{macmurchyAcceptabilityFeasibilityCost2017}.

\begin{figure}[!t]
    \begin{subfigure}[b]{0.49\textwidth}
        \centering
        \includegraphics[width=\linewidth]{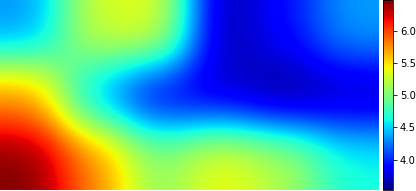}
        \caption{Accuracy result per marker aggregated over all participants}
        \label{subfig:accuracy_result_median}
    \end{subfigure}
    \begin{subfigure}[b]{0.49\textwidth}
        \centering
        \includegraphics[width=\linewidth]{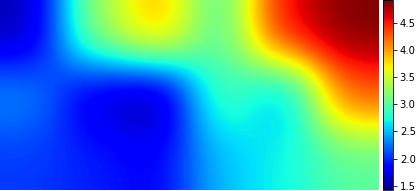}
        \caption{Precision result per marker aggregated over all participants}
    \end{subfigure}
	\caption{Aggregated results for all participants demonstrating the accuracy and precision per marker where X \& Y axes represent the ArUco marker location on screen in Pixels. The sensitivity of the camera's position relative to the viewer is evident where the highest accuracy and precision is when the user's head pose is directed at the camera.}
	\label{fig:accuracy_results_aggregated}
\end{figure}

\section{Results}
By utilising two cameras placed at a clinically safe distance, the system meets the clinical requirement of being non-invasive. \ac{RT-GENE} and \ac{RT-BENE} ensure that patient calibration is not required for the accurate and continuous measurement of gaze throughout the patient's \ac{ICU} stay.

Figure~\ref{fig:requirement_evaluation} demonstrates the distribution of required accuracy and precision measurements. The required accuracy \& precision density plot was of measured objects from the perspective of the patient in an \ac{ICU} bed-space. Pairwise distances were measured as the shortest distance between objects. Size/Distance measurements were then converted to angles from the perspective of the patient as they are sitting in an \ac{ICU} bed. This reveals that an accuracy of \requiredacc and a precision of \requiredprec is thus required for a clinically useful system.

Figure~\ref{fig:requirement_evaluation} also demonstrates the accuracy and precision results from the replicated laboratory \ac{ICU} across all participants pooled across all marker locations. Median accuracy is \evalacc with a median precision of \evalprec as defined by Equations~\ref{eq:accuracy} \& \ref{eq:precision} respectively. These results exceed empiric requirements making the system's performance suitable for use in \ac{ICU}.

In the laboratory experiment, grouping by markers, Figure~\ref{fig:accuracy_results_aggregated} displays aggregated results across all participants where each marker's median accuracy and precision are rasterised as an image the size of the marker and then smoothed with a Gaussian kernel. As the utility of this eye-tracking system is in \ac{ICU}, a median error of \evalacc is lower than the \requiredacc. The figure demonstrates the sensitivity of the camera's placement relative to the user's head as the error of the system is not uniform; median accuracy is lowest at the bottom left which, from the perspective of the head-camera, represents the most challenging eye image to regress as the eye is mostly closed. Median precision is worse in the top-right where the eye is the most open but the head is rotated giving only 1 eye to regress a gaze vector thus giving the least precision.

Optimisation of the pipeline through internal validation experiments led to acceptable performance on commercially available hardware. The total rate of 28Hz is of sufficient performance to yield to eye movement analysis.

\begin{figure*}[t]
    \centering{}
    \input{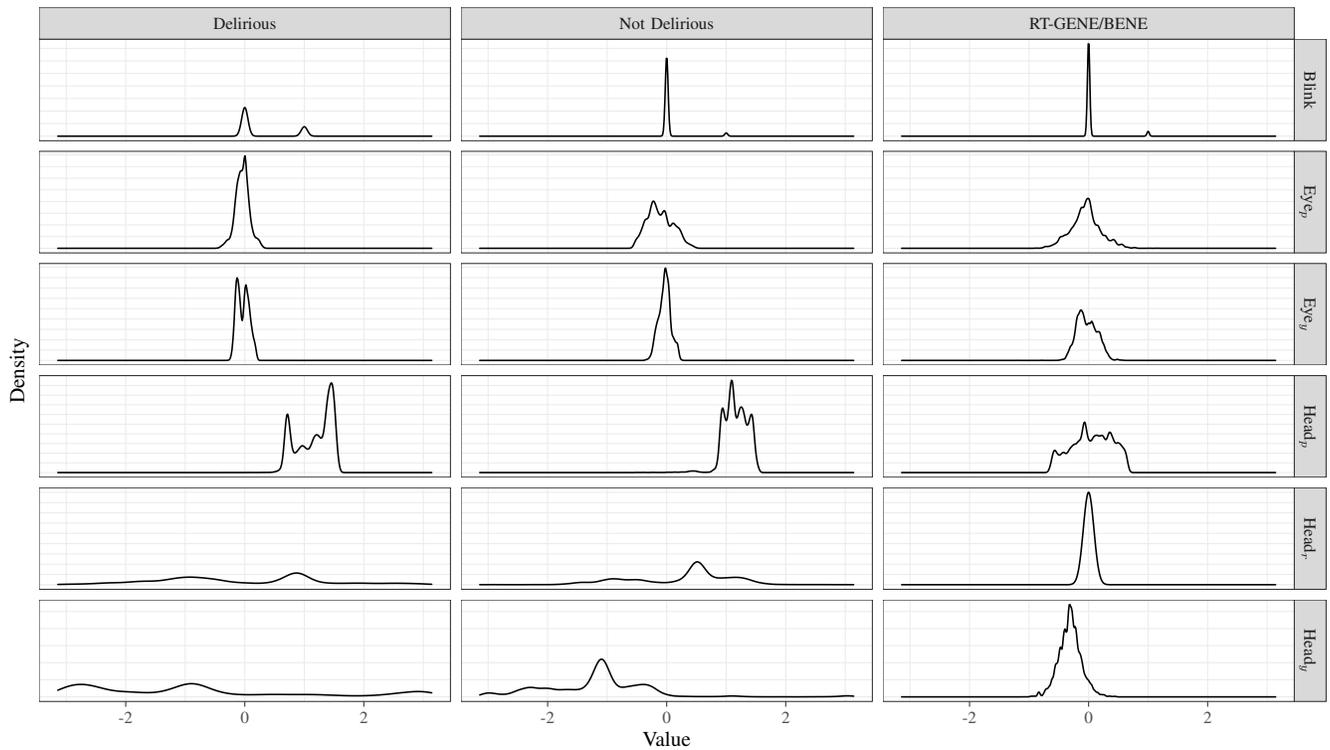}    
    \caption{Density plots of aggregated data from 5 patients. Some experienced episodes of delirium. Eye and Head refer to Eye and Head pose angles from the system while subscripts $y$, $p$ and $r$ refer to yaw, pitch and roll respectively. Blink is thresholded as per RT-BENE. Blink and Gaze statistics are of similar distributions between RT-GENE/BENE and that of non-delirious patients. Of note, no Head$_r$ data is present in \ac{RT-GENE}-\ac{RT-BENE} datasets as faces processed to remove roll prior to inference.}
    \label{fig:gaze_blink_headpose_stats}
\end{figure*}

Figure~\ref{fig:gaze_blink_headpose_stats} demonstrates the measurements that the system generates on patients in \ac{ICU}. The system was successfully deployed on 5 patients where some of whom developed episodes of delirium. Blinks, eye and head angles are displayed as aggregated density plots. Extreme head angles are as a result of occlusions of the patient's face - examples of such occlusions are the patient's hands occluding their face, nurses interrupting the line of sight between the camera and the patient and nasogastric tubes. These periods of occlusions could be mediated by increasing the threshold for face detection above the current value of 0.6 thus increasing the specificity of detections ignoring periods of occlusion.

The figure also demonstrates the aggregate statistics from the datasets used to create \ac{RT-GENE} and \ac{RT-BENE} alongside. The distributions of blinks and gaze data are similar across the patients and the dataset indicating likely accurate acquisition.

The questionnaire outlined in Table~\ref{tab:nurse_questionnaire} was conducted on 11 nursing staff, all of whom reported "No" to every question indicating the acceptability of the system in clinical practice.



\begin{table}[b]
    \centering
    \begin{tabular}{p{0.8\linewidth}p{0.1\linewidth}}
    \toprule
    Question                       & Answer \\ \midrule
    Did this study affect your ability to care for your patients? & Yes/No  \\
    Did this study affect your interactions with patients/families? & Yes/No \\
    Did this study affect your interactions with other nurses? & Yes/No \\
    Did this study affect your interactions with doctors? & Yes/No \\ \bottomrule
    \end{tabular}
    \caption{This questionnaire used assess the acceptability of the use of cameras within this study
    on day to day activities of the nursing staff on ICU.}
    \label{tab:nurse_questionnaire}
\end{table}

\section{Conclusion}
Delirium is a common occurrence in \ac{ICU} with tests that are laborious or too restricted. Eye-tracking has been suggested to diagnose \textit{inattention} as a surrogate marker of delirium but to date, no system has met the requirements to be suitable for use in \ac{ICU}.
We thus developed a system that meets empiric and measured criteria as being non-invasive, calibration-free and exceeds the required accuracy and precision all while running on commercially available hardware. 

Future work will focus on the use of the signals acquired by this system to automate the diagnosis of delirium.


\bibliographystyle{./IEEEtrans} 
\bibliography{zotero_lib}

\end{document}